\documentclass[letterpaper, 10 pt, conference]{ieeeconf}  
\usepackage[utf8]{inputenc}
\usepackage{subcaption}    
\usepackage{mathtools}
\usepackage{amsmath}
\usepackage{float}
\usepackage{siunitx}
\usepackage{amssymb }
\usepackage{gensymb}
\usepackage{graphicx}
\usepackage{siunitx}
\DeclareMathAlphabet{\mathcal}{OMS}{cmsy}{m}{n}
\DeclareSymbolFont{largesymbols}{OMX}{cmex}{m}{n}

\IEEEoverridecommandlockouts                              

\usepackage{epsfig} 
\usepackage{mathptmx} 
\usepackage{times} 
\usepackage{amsmath} 
\usepackage{amssymb}  
\usepackage{graphicx} 
\usepackage{url}

\title{\LARGE \bf
	LPVIMO-SAM: Tightly-coupled LiDAR/Polarization Vision/Inertial/Magnetometer/Optical Flow Odometry via Smoothing and Mapping
}

\author{Derui Shan$^{1}$, Peng Guo$^{1}$, Wenshuo Li$^{2}$, Du Tao$^{1}$*. 
	\thanks{This work is supported by National Key Research and Development Program of China under Grant 2023YFB4704404, in part by National Natural Science Foundation of China No. 62388101. \newline
		\hspace*{1.25em}$^{1}$D. Shan, P. Guo and D. Tao are with the Faculty of Information, North China University of Technology, Beijing, China. \newline
		\hspace*{1.25em}$^{2}$W. Li are with the Hangzhou Innovation Institute, Beihang University, Hangzhou, China.}
}

\begin{document}

	\maketitle
	\thispagestyle{empty}
	\pagestyle{empty}

	\begin{abstract}
	We propose a tightly-coupled LiDAR/Polarization Vision/Inertial/Magnetometer/Optical Flow Odometry via Smoothing and Mapping (LPVIMO-SAM) framework, which integrates LiDAR, polarization vision, inertial measurement unit, magnetometer, and optical flow in a tightly-coupled fusion. This framework enables high-precision and highly robust real-time state estimation and map construction in challenging environments, such as LiDAR-degraded, low-texture regions, and feature-scarce areas. The LPVIMO-SAM comprises two subsystems: a Polarized Vision-Inertial System and a LiDAR/Inertial/Magnetometer/Optical Flow System. The polarized vision enhances the robustness of the Visual/Inertial odometry in low-feature and low-texture scenarios by extracting the polarization information of the scene. The magnetometer acquires the heading angle, and the optical flow obtains the speed and height to reduce the accumulated error.  A magnetometer heading prior factor, an optical flow speed observation factor, and a height observation factor are designed to eliminate the cumulative errors of the LiDAR/Inertial odometry through factor graph optimization. Meanwhile, the LPVIMO-SAM can maintain stable positioning even when one of the two subsystems fails, further expanding its applicability in LiDAR-degraded, low-texture, and low-feature environments. Code is available on https://github.com/junxiaofanchen/LPVIMO-SAM.
	\end{abstract}

	\section{INTRODUCTION}
	Simultaneous Localization and Mapping (SLAM) is a method for localization and map reconstruction  simultaneously in unknown environments. Over the past two decades, great  efforts have been made in implementing SLAM in unknown environments using a single sensor, such as LiDAR \cite{zhang2014loam}-\cite{shan2018lego} or cameras \cite{campos2021orbslam3}. 
	LiDAR-based methods can capture detailed environmental information from long distances. However, they may fail to operate when the environmental structural information is sparse or missing. Visual-based methods are suitable for position recognition, but their accuracy often drops significantly in low-texture environments. To this end, multi-sensor fusion has attracted increasing attention \cite{qin2018vinsmono}-\cite{shan2020liosam}. Using  optimization methods to make multi-sensors integrate to accurate pose estimation in sensor-degraded environments.
	
	In the existing SLAM frameworks, the most widely used ones are cameras, LiDARs, and inertial measurement unit (IMU). In recent years, various LiDAR-Visual-Inertial odometry (LIVO) systems have been proposed to achieve highly robust and accurate state estimation, such as R2LIVE \cite{lin2021r2live} and LVI-SAM \cite{shan2021lvisam}. These LIVO systems typically consist of a LiDAR/Inertial odometry (LIO) subsystem and a Visual/Inertial odometry (VIO) subsystem, which first process each type of data separately and then jointly fuse the state estimation. However, these subsystems fail to address the issues like the Z-axis drift  of the LIO system in unstructured scenarios \cite{xu2023golio}, the problem that low pose estimation accuraccy of the VIO system in low-texture scenarios \cite{wang2024lowtexture}. To address these issues, we introduces a tightly-coupled LiDAR/Polarization Vision/Inertial/Magnetometer/ Optical Flow odometry (LPVIMO-SAM) via smoothing and mapping . By the complementary of multiple sensors, the proposed method can realize more robust and accurate pose estimation. The contributions of this paper are as follows: 
	
	1) To overcome the navigation and positioning challenges in low-texture and LiDAR-degraded environments, a tightly-coupled LPVIMO-SAM framework based on factor graphs is proposed. This system integrates LiDAR, polarization vision, IMU, magnetometer and optical flow ranging module.
	
	2) For the issues of Z-axis drift and heading error accumulation from the LIO subsystem in LiDAR-degraded or unstructured environments, an optical flow ranging module is introduced to construct height factors and velocity factors, which reduce Z-axis drift in the LIO system. Additionally, a magnetometer sensor is used to provide the initial heading and serve as a prior factor to mitigate the accumulation of heading errors.
	
	3) For the problems such as feature extraction and matching failure of the VIO subsystem in low-texture scenes, an polarization camera is introduced into the tightly-coupled system. The degree of polarization (DOP) and angle of polarization (AOP) images acquired by the polarization camera are used to assist the feature extraction of grayscale images, enhancing the feature-extraction capability of the VIO subsystem in low-texture scenes. Then, the enhanced VIO subsystem is integrated with the LIO subsystem.

	\section{RELATED WORKS}
	We divide the work related to the system proposed in this paper and the involved technologies into the following two parts:
	\subsection{LiDAR-Visual-Inertial SLAM}
	Fusing multiple sensors in LVIS SLAM enables the system to achieve robust localization in challenging environments, especially when one of the sensors fails or its performance degrades. Recently, various LiDAR-Visual-Inertial SLAM systems have emerged in the research community. According to whether the LIO system and the VIO system are fused at the state estimation level and the raw measurement level, LIVO systems can be divided into loosely-coupled and tightly-coupled categories.
	
	VIL-SLAM introduces a LiDAR-Visual-Inertial system (LVIS) that is loosely-coupled at the state estimation level. The LIO subsystem only uses the initial pose provided by the VIO subsystem, and there is no joint optimization of the measurement values of LiDAR, cameras, and IMU \cite{shao2019stereovilidar}. Meanwhile, some systems \cite{zhu2021camvox}-\cite{huang2020lidarmonocular} use 3D LiDAR point clouds to provide depth measurements for the visual module \cite{campos2021orbslam3}-\cite{qin2018vinsmono}. Since these systems don't directly obtain constraints from LiDAR measurements in state estimation, they remain loosely-coupled. Based on this, Cheng et al. introduced the keyframe and sliding window algorithm as well as the loop detection algorithm based on iterative closest point at the backend, reducing the computational load and large-scale drift. Compared with ORB-SLAM3, the Root Mean Square Error (RMSE) decreased by approximately 70\% \cite{cheng2024vehiclelocalization}.  To improve accuracy and robustness, the approach of tightly-coupled joint optimization of sensor data has been continuously updated and refined. Based on the MSCKF framework \cite{mourikis2007msckf}, LIC-Fusion tightly integrates IMU data, sparse visual features, and planar and edge features of LiDAR. In outdoor scenarios, its absolute trajectory error is approximately 62\% lower than that of the MSCKF method \cite{zuo2019licfusion}.
	
	In recent years, an increasing number of systems have achieved full tightly-coupled in  measurement and state estimation. LVI-SAM bases on the factor graph framework to tightly-couple LiDAR, visual, and inertial sensors, with its VIO subsystem using LiDAR scans to track visual features and extract feature depth, and compared with the LIO-SAM system, the RMSE is reduced by approximately 81\% \cite{shan2021lvisam}. FAST-LIVO presents a fast, sparse-direct, image-to-map the LVI fusion framework, achieving multi-sensor fusion through error state iterative Kalman Filtering, and compared with R2LIVE, its RMSE is reduced by approximately 58\% \cite{zheng2022fastlivo}. Subsequently, FAST-LIVO2 is proposed to visual tracking at the image patch level, and reduced the cost of residual construction and rendering. Compared with the FAST-LIVO system, its RMSE decreased by approximately 50\% \cite{zheng2025fastlivo2}. LVIO-Fusion's VIO subsystem obtains real radiometric information through online photometric calibration, capturing subtle geometric and radiometric changes. This effectively overcomes sensor degradation issues in environments with no geometry and no texture, enhancing the system's robustness. The positioning accuracy is improved by approximately 47.8\% compared to FAST-LIVO \cite{zhang2024lviofusion}. However, research on existing methods in low-texture and LiDAR-degraded scenarios remains limited. The LIVO system still faces issues such as scan matching failures in the LIO subsystem and feature extraction and matching failures in the VIO subsystem. These issues still require further research and solutions.
	
	\subsection{Polarization Integrated Navigation System}
	Many animals, such as desert ants and mantis shrimps, can sense the polarization light, which helps them achieve accurate navigation and orientation through polarized visual \cite{dupeyroux2019antbot}-\cite{chiou2020neural}. In recent years, biomimetic polarized light sensors have been widely used in navigation because they can obtain precise heading angle from the polarized skylight or scene \cite{wang2015positionmethod}.
	
	From the source of  polarization information, it can be categorized  into sky polarization and scene polarization.
	Most existing integrated navigation methods rely on making fully use of sky polarization.
	A bionic polarized attitude and heading reference system was proposed to improve  the accuracy of attitude of inertial sensors, the standard deviations of the pitch and roll  were reduce to \SI{0.1}{\degree}, and  the heading  was \SI{0.90}{\degree} \cite{yang2020polarizationahrs}. A bionic polarized light sensor/INS/wheel odometer integrated navigation system  was developed in \cite{dou2022polarizednavigation}, and a marginalized Unscented Kalman Filter was utilized to reduce the RMSE to  approximately  \SI{1.70}{\degree} in a hexagonal trajectory. 
	The factor graph optimization method  was introduced  to obtain the optimal solution of  INS/LiDAR/Polarized camera integrated  navigation system to improve the pose estimation accuracy \cite{du2022inslidar}. To eliminate the  multi-source disturbance from  inaccurate modeling and unknown parameters  of  Polarized/Inertial/GNSS integrated navigation systems, an outlier-robust extended Kalman filter were designed  to  increase accuracy by  about 50\% compared with GNSS/INS integrated navigation systems \cite{qiu2024robustekf}. A tightly-coupled Polarization Vision/Inertial positioning system was developed to heading determination and positioning \cite{wan2024visualinertial}. Subsequently, polarization  skylight sensors were conducted to  successful application in underwater and moonlight  environments  \cite{zhao2024solartracking}-\cite{chen2024autonomouspositioning}.

	For the scene polarization,  it was also used to  target detection \cite{wang2017bionicorientation},  dense reconstruction \cite{yang2018polarimetricslam}, and  enhance performance of ORB-SLAM3 \cite{Du2025ORBSLAM}. However, the scene polarization is rarely used to LiDAR/Visual/Inertial fusion. Therefore, we uses the degree of polarization (DoP) images and angle of polarization (AoP) images  to  feature extraction.
	The scene polarization can be used in low-texture and low-feature scenes, and  enhancing the robustness of the VIO system. Consequently, a highly robust LiDAR/IMU/Polarization Vision/Magnetometer/Optical Flow system is obtained.
	
	\begin{figure*}[htbp!]
		\centering
		\vspace*{5pt} 
		\includegraphics[width=15cm]{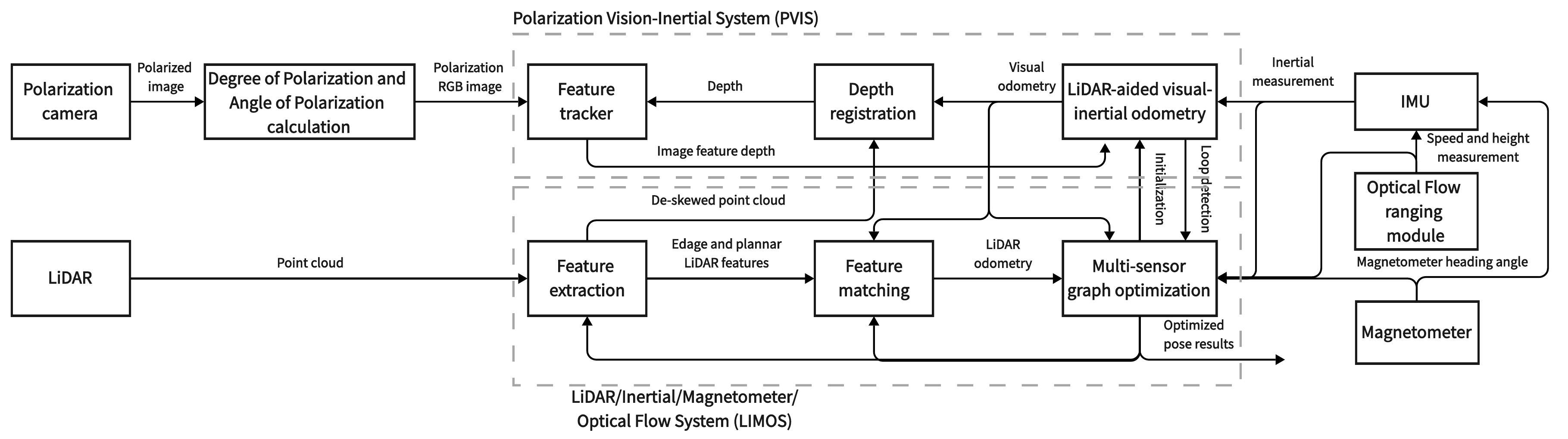}
		\caption{Flow chart of the LPVIMO-SAM system.}
		\label{fig1}
	\end{figure*}
	
	\section{LIDAR/POLARIZATION VISION/INERTIAL/ MAGNETOMETER/OPTICAL FLOW RANGING MODULE ODOMETRY VIA SMOOTHING AND MAPPING}
	
	\subsection{System Overview}
	As shown in Fig. 1, the LiDAR/Polarization Vision/Inertial /Magnetometer/Optical Flow system (LPVIMOS) consists of  a LiDAR, a polarization camera, an IMU, a magnetometer sensor, and an optical flow ranging module, and it is composed of two subsystems: a Polarization Vision/Inertial System (PVIS) and a LiDAR/Inertial/Magnetometer/Optical Flow System (LIMOS). PVIS processes polarized RGB images and IMU measurements, and utilizes LiDAR data to obtain the depth of visual features. For the specific establishment process and principles of visual/Inertial System (VIS), please refer to \cite{qin2018vinsmono}.
	LIMOS extracts LiDAR features and obtains the LiDAR odometry by matching the extracted feature with the feature map. We used a sliding window to maintain the feature map to ensure real-time performance. Finally, we jointly optimized the contributions of various sensors within a factor graph using iSAM2 \cite{kaess2012isam2}.
	\par
	These contributions include IMU pre-integration constraints, polarization vision/inertial odometry constraints, magnetometer heading constraints, LiDAR odometry constraints, velocity and height constraints from the optical flow odometry module, and loop closure constraints. The state estimation  is formulated as a maximum a posterior estimation. Note that in this paper, the RGB images from the polarization camera are used to improve the robustness of the VIS, and the optical flow and the magnetometer sensor are used to improve the positioning accuracy of the integrated navigation system and the robustness of the LIS.
	
	\subsection{Calculation of Polarized RGB Images}
	\begin{figure}[htbp]
		\centering
		\includegraphics[width=4cm]{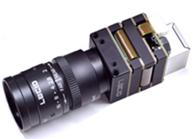}
		\includegraphics[width=3cm]{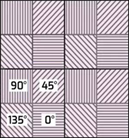}
		\caption{Left: Polarized image sensor PHX050S-PC, Right: Pixel distribution map.}\label{fig2}
	\end{figure}
	The grayscale image, DOP image, and AOP image are calculated using the following formulas respectively:
	The polarized camera sensor  (Sony IMX250MZR CMOS) are used to acquires raw polarization images of size $2448\times2048$ as single-channel grayscale images as shown in Fig. 2. The  frequency is 20 Hz. Four pixels form a unit, and the polarization images can be captured simultaneously in four different directions $\left(0,\, \pi/4,\, \pi/2,\, 3\pi/4 \right)$. 
	
	1)	grayscale image
	
	A new pixel is calculated for each unit of the original image. The grayscale value g of this pixel can be obtained as follows:
	\begin{equation}
		\begin{aligned}
			g=\frac{1}{2}\left(g_{0}+g_{1}+g_{2}+g_{3}\right), \quad g \in (0,255)
		\end{aligned}
	\end{equation}
	$g_{0}, \ g_{1}, \ g_{2}$ and $g_{3}$ represent the grayscale values produced by light passing through polarizers oriented at \SI{0}{\degree}, \SI{45}{\degree}, \SI{90}{\degree} and \SI{135}{\degree}, respectively. The grayscale images with a resolution of $1224\times1024$ pixels.
	
	2)	DOP image
	
	For each unit in the original image, a new pixel is calculated, where the degree of polarization of this pixel is computed by Eq. (2).
	\begin{equation}
		\begin{aligned}
			d=2 \frac{\sqrt{\left(I_{0^{\circ}}-I_{90^{\circ}}\right)^{2}+\left(I_{45^{\circ}}-I_{135^{\circ}}\right)^{2}}}{I_{0^{\circ}}+I_{45^{\circ}}+I_{90^{\circ}}+I_{135^{\circ}}}
		\end{aligned}
	\end{equation}
	$I_{0^{\circ}}, \ I_{45^{\circ}},  \  I_{90^{\circ}}$ and $I_{135^{\circ}}$ represent the intensities of scene light waves after passing through polarizers oriented at \SI{0}{\degree}, \SI{45}{\degree}, \SI{90}{\degree} and \SI{135}{\degree}, respectively. The DOP d  is nonlinearly mapped to an integer between 0 and 255 to enhance the contrast of the DOP image.
	\begin{equation}
		\begin{aligned}
			g_{d}=-255 d^{2}+510 d, d \in[0,1]
		\end{aligned}
	\end{equation}
	The grayscale value of each pixel is replaced by the $g_{d}$ value after polarization degree mapping, resulting in the DOP image.
	
	3)	AOP image
	
	For each unit in the original image, a new pixel is calculated, where the polarization angle of this pixel is computed by Eq. (4)-(5). The AOP $\theta$ is then linearly mapped to an integer between 0 and 255.
	\begin{equation}
		\begin{aligned}
			\theta=\frac{1}{2} \arctan \left(\frac{I_{45^{\circ}}-I_{135^{\circ}}}{I_{0^{\circ}}-I_{90^{\circ}}}\right)
		\end{aligned}
	\end{equation}
	\begin{equation}
		\begin{aligned}
			g_{\theta}=\left(\theta+\frac{1}{2} \pi\right) * \frac{255}{\pi}, \theta \in\left[-\frac{\pi}{2}, \frac{\pi}{2}\right]
		\end{aligned}
	\end{equation}
	The grayscale value of each pixel is replaced by the $g_{\theta}$ value after the polarization angle mapping, resulting in the AOP image.
	
	After calculating the grayscale image, DOP image, and AOP image from a single original image, these three single-channel images are combined into an RGB image. In this image, the Red channel represents the polarization degree image, the Green channel represents the grayscale image, and the Blue channel represents the polarization angle image.
	\subsection{Polarization Vision/Inertial System}
	The LPVIMO-SAM system  is inspired by  the LVI-SAM  system framework.
	We introduces polarized RGB images and utilizes the  DOP and AOP images to enhance the feature of grayscale images.
	In complex scenes with low-texture and low-feature, where ordinary VIS systems may lose feature or fail to match, the  proposed PVIS leverages scene polarization  to supplement feature information. This enables the VIS system to stably and efficiently extract and track features even in complex environments. Due to space limitations, the design of VIS follows the approach in \cite{shan2021lvisam}.
	\begin{figure}[htbp!]
		\centering
		\includegraphics[width=8cm]{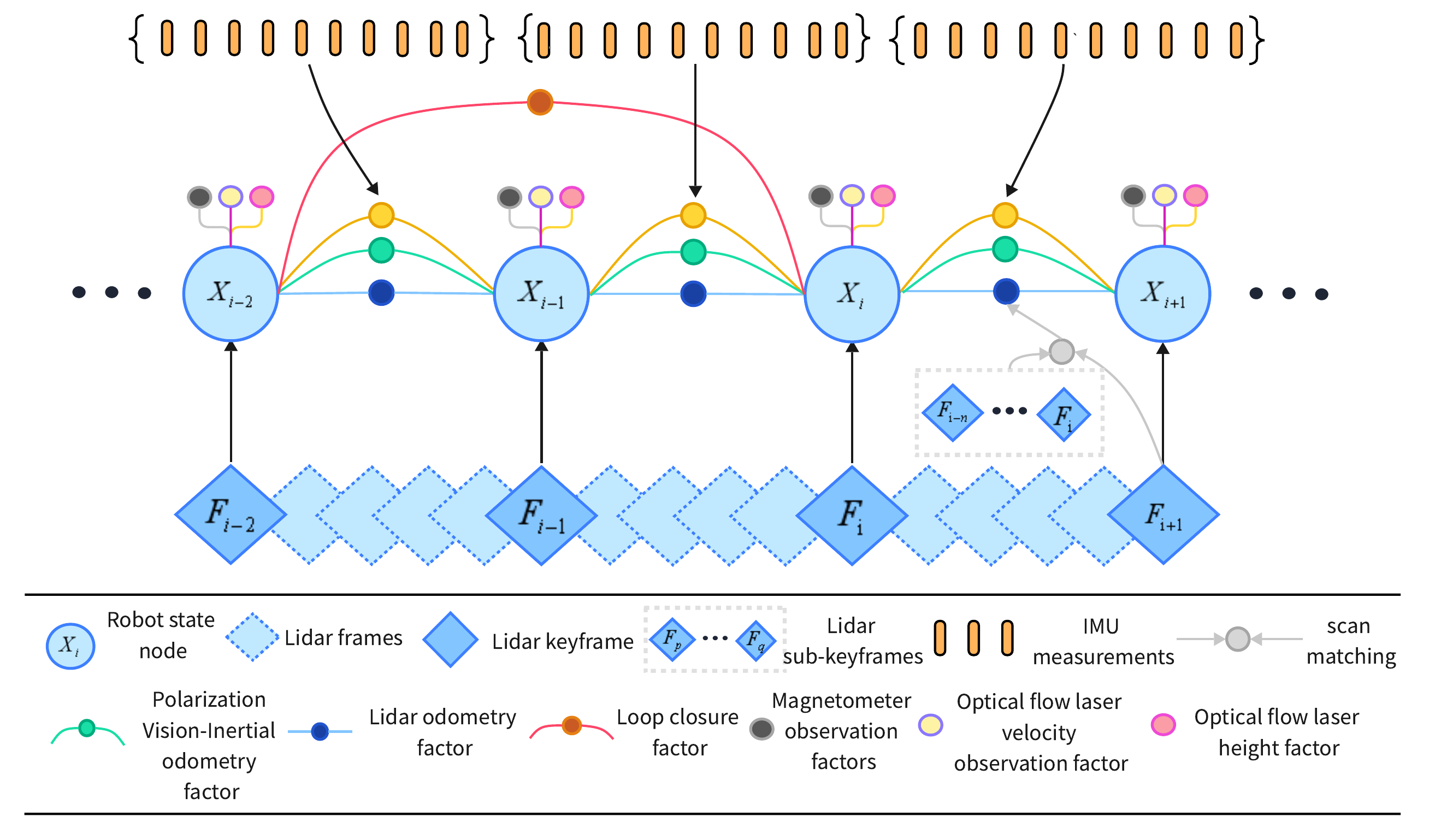}
		\caption{The factor graph framework of the LiDAR/Polari-zation Vision/Inertial/Magnetometer/Optical Flow navigation system (adaptive modified from \cite{shan2021lvisam}).}
		\label{fig3} 
	\end{figure}
	
	\subsection{LiDAR/Inertial/Magnetometer/Optical Flow System}
	The LIS system in the LPVIMO-SAM framework proposed in this paper is illustrated in Fig. 3. Building upon the LIS system in LVI-SAM, magnetometer heading factors are introduced to correct the heading error, while velocity factors and height factors derived from the optical flow ranging module are used to constrain the system's velocity and Z-axis error. Compared to the factor graph in the PFLIO-SAM \cite{Shan2025pfliosam}, the factor graph in this article replaces the Polarization Heading factor constraint with a magnetometer heading factor constraint, and additionally adds a Polarization Vision-Inertial Odometry factor constraint.
	
	1)	 IMU pre-integration factor and LiDAR odometry factor: The LIMOS system  is built upon the LIS system in LVI-SAM. Therefore, the design of these factors and the system can be referred to \cite{shan2020liosam}.
	
	2)	Magnetometer heading factor: A magnetometer is introduced to obtain the absolute heading  of the vehicle in the world coordinate system, thereby enabling the heading angle of the navigation and positioning system to align directly with the world coordinate system. Subsequently, the difference between the magnetometer heading angle and the optimized heading angle from the system is calculated, and a threshold is set to filter out invalid measurements of the magnetometer heading angle. Finally, the magnetometer heading angle is used as a heading observation factor and inserted into the factor graph.
	
	3)	Velocity factor and Height factor of the optical flow: Due to space limitations, the design of this part can refer to the factor design in the PFLIO-SAM paper \cite{Shan2025pfliosam}.
	\section{EXPERIMENTS}
	A series of experiments on the proposed framework were conducted to obtain five self-gathered datasets, which are named scenes a to e. The sensor suite for data collection includes a Velodyne VLP16 LiDAR, an IMX250MZR CMOS camera, a WTGAHRS1 IMU, an optical flow ranging module, and a JY-RM3100 magnetometer. The trajectory ground truth is provided by RTK (Planar error: \SI{0.008}{\metre}, Elevation error: \SI{0.015}{\metre}). We compare the proposed framework with several open-source solutions, including LIO-SAM, FAST-LIO2, LVI-SAM, and FAST-LIVO.
	
	The five datasets  were collected in the playground, small basketball court, volleyball court, and large basketball court below the Boyuan Building at the North China University of Technology.
	Their respective characteristics are as follows:
	
	1) Scene a: The small basketball court with abundant Zigzag trajectories.
	
	2)  Scene b: The small basketball court with a high number of dynamic pedestrians.
	
	3)	Scene c: The large basketball court, with an LiDAR-degraded environment and sparse features. 
	
	4)	Scene d: The combination of the small basketball court and the volleyball court feature a higher number of dynamic pedestrians.
	
	5)	Scene e: Playground scene, with an LiDAR-degraded environment and sparse feature.

	\begin{table}[htbp!]
		\centering
		\setlength{\tabcolsep}{0.8pt} 
		\renewcommand{\arraystretch}{1.3} 
		\caption{Comparison of trajectory accuracy (RMSE/\si{\meter}) among different algorithms.}
		\label{tab1}
		\begin{tabular}{cccccccc}
			\hline
			\begin{tabular}[c]{@{}c@{}}Scene\\ name\end{tabular}       & \begin{tabular}[c]{@{}c@{}}LIO-\\ SAM\end{tabular} & \begin{tabular}[c]{@{}c@{}}FAST-\\ LIO2\end{tabular} & \begin{tabular}[c]{@{}c@{}}LIOMO-\\ SAM\end{tabular} & \begin{tabular}[c]{@{}c@{}}LVI-\\ SAM\end{tabular} & \begin{tabular}[c]{@{}c@{}}FAST-\\ LIVO\end{tabular} & \begin{tabular}[c]{@{}c@{}}LVI-\\ SAM(Harsh)\end{tabular} & \begin{tabular}[c]{@{}c@{}}LPVIMO-\\ SAM(Harsh)\end{tabular} \\ \hline
			\begin{tabular}[c]{@{}c@{}}Scene-a\\ (\SI{165.9}{\metre})\end{tabular} & 1.66                                               & 1.15                                                 & 0.57                                                 & 1.66                                               & 3.08                                                 & Fail                                                      & 0.54                                                         \\
			\begin{tabular}[c]{@{}c@{}}Scene-b\\ (\SI{143.7}{\metre})\end{tabular} & 1.79                                               & 1.51                                                 & 0.77                                                 & 1                                                  & 3.99                                                 & Fail                                                      & 0.74                                                         \\
			\begin{tabular}[c]{@{}c@{}}Scene-c\\ (\SI{232.5}{\metre})\end{tabular} & 2.27                                               & 1.01                                                 & 0.86                                                 & 1.91                                               & 1.31                                                 & Fail                                                      & 0.83                                                         \\
			\begin{tabular}[c]{@{}c@{}}Scene-d\\ (\SI{256.1}{\metre})\end{tabular} & 2.98                                               & 1.39                                                 & 0.91                                                 & 1.62                                               & 5.52                                                 & 9.08                                                      & 0.87                                                         \\
			\begin{tabular}[c]{@{}c@{}}Scene-e\\ (\SI{480.3}{\metre})\end{tabular} & Fail                                               & 3.78                                                 & Fail                                                 & 2.88                                               & 5.03                                                 & 13.18                                                     & 2.28                                                         \\ \hline
		\end{tabular}
	\end{table}
	\subsection{Comparison of LiDAR-Inertial Systems}
	The datasets from five scenarios with different lengths are used to verify the effectiveness of the proposed method. In scene a,  more zigzag trajectories are introduced. In this case, both the LIO-SAM and FAST-LIO2 systems demonstrate good positioning performance. However, with the constraints of the magnetometer heading and the velocity and height from the optical flow ranging module, the proposed LIOMO-SAM improves the trajectory accuracy by approximately 65.7\% compared to LIO-SAM.
	
	In scene b, c, and d, the trajectory length was increased. Due to error accumulation, the LIO-SAM system suffered from a severe drift problem in the Z-axis. As the trajectory length of the LIO-SAM system increased, its RMSE continued to rise. In contrast, the trajectory positioning accuracy of FAST-LIO2 remained relatively stable. The proposed LIOMO-SAM constrains the Z-axis error and velocity error by using the optical flow. Consequently, its average trajectory positioning accuracy is approximately 62.9\% higher than that of LIO-SAM.
	
	In scene e, affected by the LiDAR-degraded environment, the LiDAR scan matching of the LIO-SAM system failed, causing the carrier's positioning to either stay in place continuously or experience severe drift. The LIOMO-SAM corrected the entire system by adding constraints, enabling the system to operate effectively for a period of time. However, as errors continued to accumulate, the impact of the LiDAR odometry failure became more prominent, and ultimately, the LIOMO-SAM system also experienced significant drift and failed to complete the trajectory positioning. In contrast, FAST-LIO2 can still achieve positioning but with relatively large RMSE values.
	
	\subsection{Comparison of LiDAR/Inertial/Visual fusion Systems}
	To further enhance the system's reliability, the VIS is introduced into the LIS. This addition provides more constraints for the LIS, enabling the entire system to operate stably and achieve accurate positioning even when the LiDAR scan-matching fails. For the different application scenarios, when the LVI-SAM system that combines the VIS and LIS is introduced, its positioning accuracy surpasses that of the LIO-SAM system in each scene. Obviously, in scene e where the LIO-SAM system fails, the LVI-SAM system can maintain stable operation and continuously output reliable positioning information. From these comparison results, the LVI-SAM system has more advantages than the LIO-SAM system in terms of stability and reliability. Since the algorithm adopted by the FAST-LIVO system provides polarized RGB images, this system experiences positioning drift in some scenarios. However, in scene c, the FAST-LIVO system can operate stably, and its positioning accuracy is slightly higher than that of the LVI-SAM system, demonstrating its advantages. Note that all subsequent mentions of LPVIMO-SAM  refer to the results under Harsh conditions. Harsh indicates a corner selection quality level of 0.9-0.95.
	\begin{figure}[htbp!]
		\centering
		\begin{subfigure}{4.2cm}
			\includegraphics[width=\textwidth]{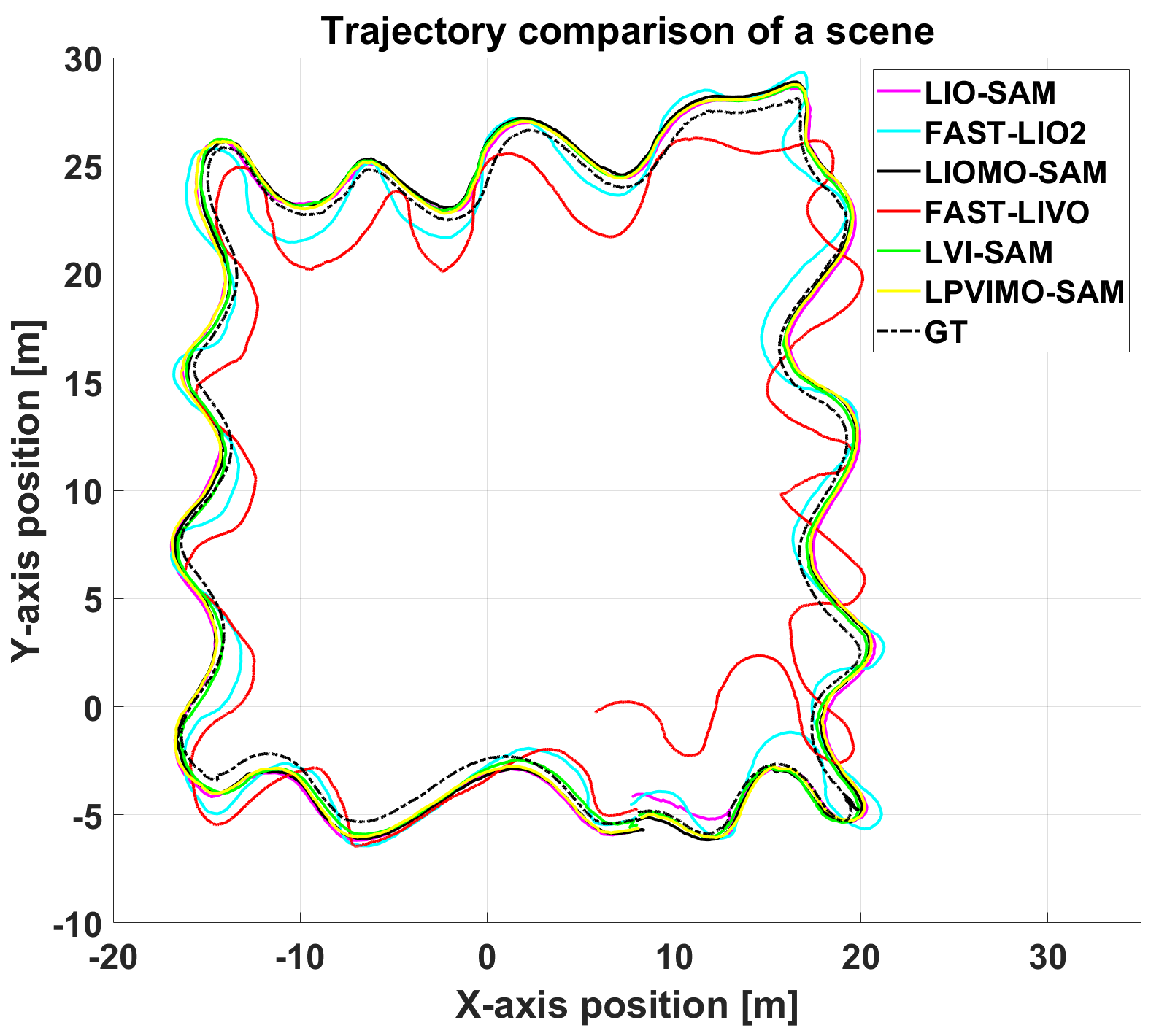}
			\caption{Scene a}\label{scene:1}
		\end{subfigure}
		\hfill
		\begin{subfigure}{4.2cm}
			\includegraphics[width=\textwidth]{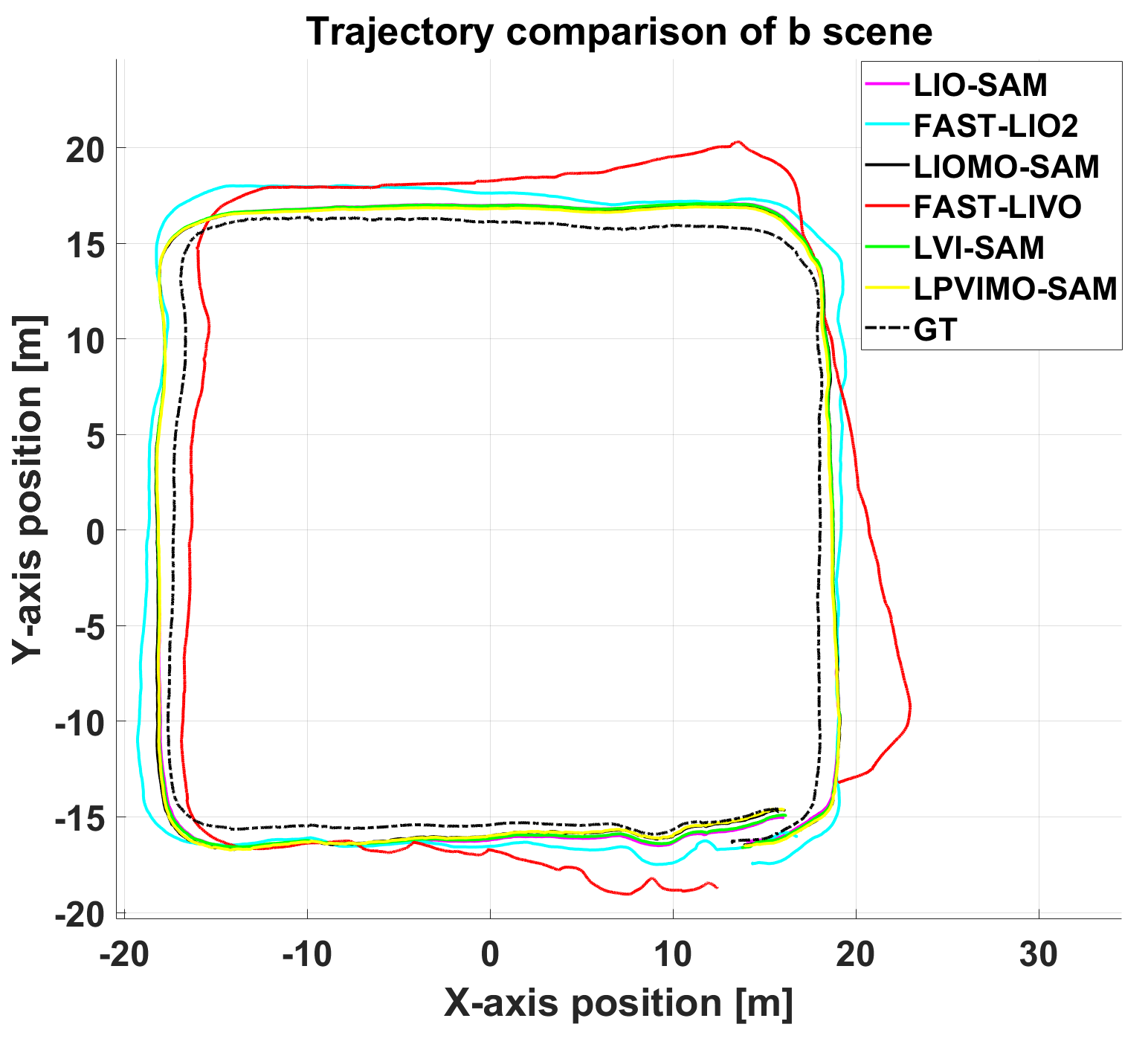}
			\caption{Scene b}\label{scene:2}
		\end{subfigure}
		
		\begin{subfigure}{8cm}
			\includegraphics[width=\textwidth]{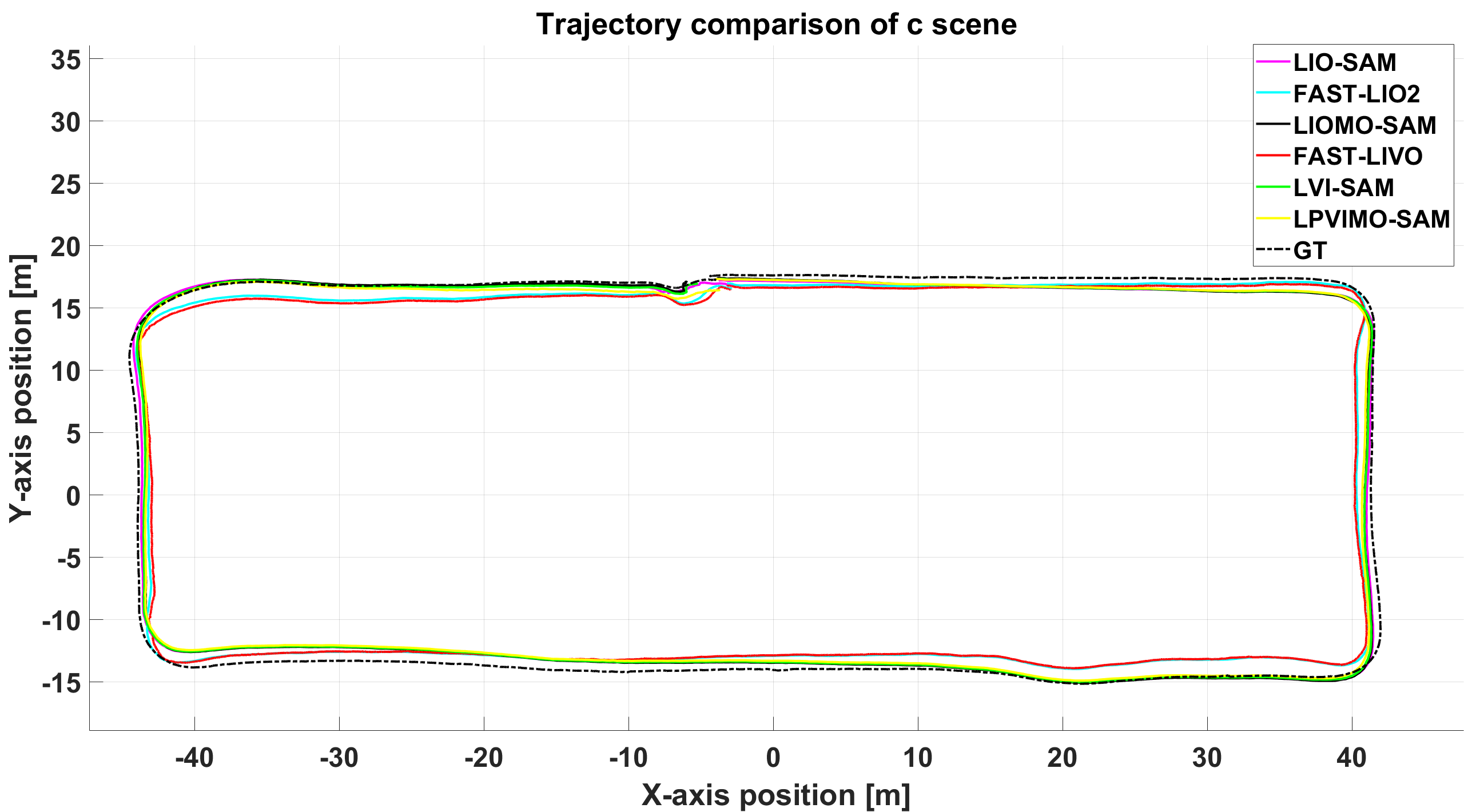}
			\caption{Scene c}\label{scene:3}
		\end{subfigure}
		
		\begin{subfigure}{4.2cm}
			\includegraphics[width=\textwidth]{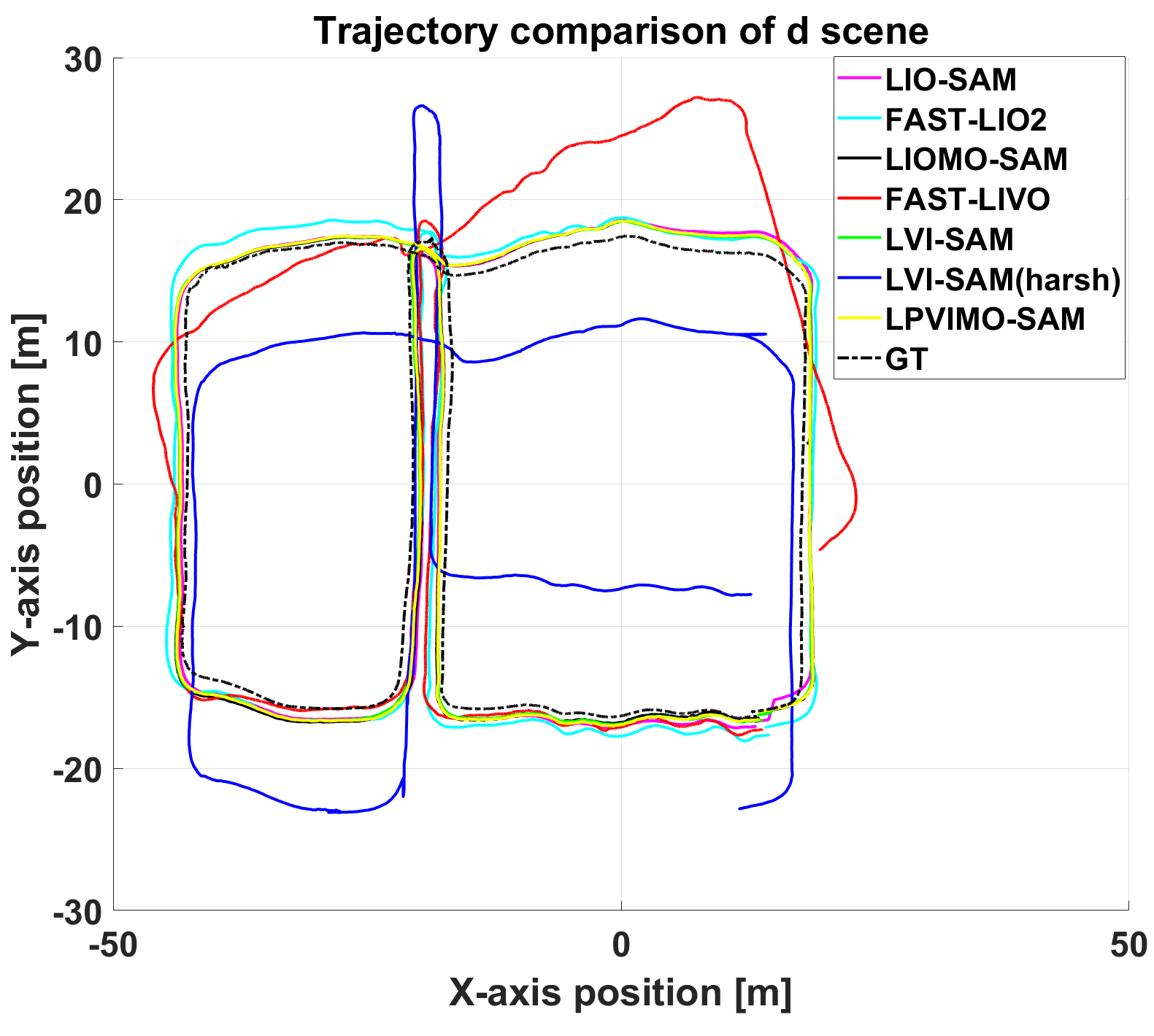}
			\caption{Scene d}\label{scene:4}
		\end{subfigure}
		\hfill
		\begin{subfigure}{4.2cm}
			\includegraphics[width=\textwidth]{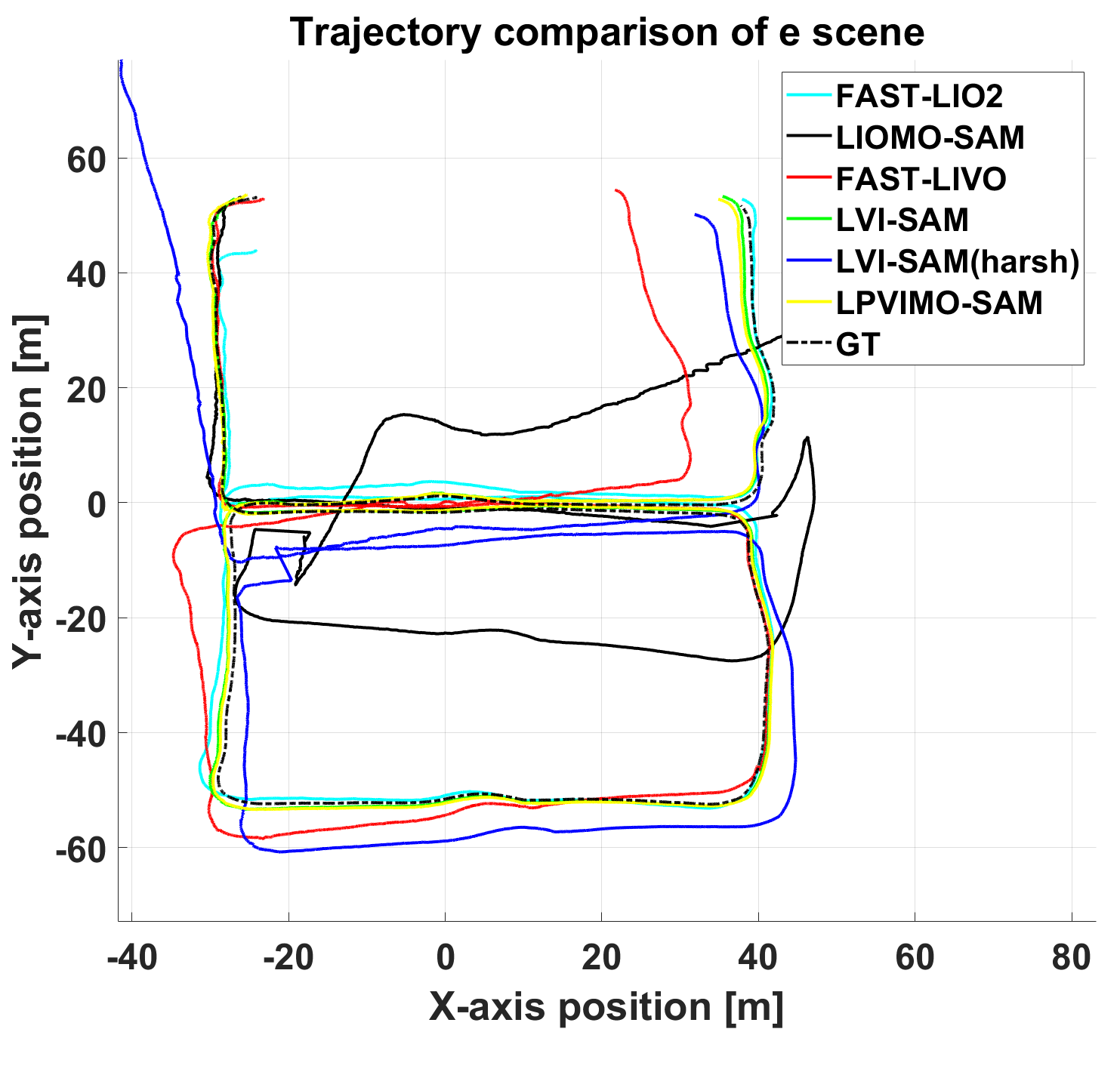}
			\caption{Scene e}\label{scene:5}
		\end{subfigure}
		
		\caption{Trajectory comparison charts for each scene.}\label{fig4}
	\end{figure}

	Fig. 4 displays trajectory for different scenarios. Scenarios a and b compare the robustness of various algorithms with zigzag trajectories. It shows that the proposed LPVIMO-SAM method exhibits excellent stability in these scenarios. Scenario c extends the path length based on scenario b, and all algorithms maintain good performance. Although scenarios c and d have similar path lengths, the trajectory in scenario d is slightly more complex, which causes the FAST-LIVO system to drift during subsequent operations. The LVI-SAM system, while capable of running completely under harsh conditions, also experiences significant drift in trajectory positioning.	Scenario d selects a significantly more LiDAR-degraded environment with a longer running distance on a playground, where all methods produce considerable errors. FAST-LIVO begins to drift at the coordinate (-40, 0). FAST-LIO2 can operate stably, but its starting node is located between X-axis \SI{-30}{\meter} to \SI{-20}{\meter} and Y-axis \SI{40}{\meter} to \SI{50}{\meter}, indicating that its error is larger than that calculated by EVO. The LVI-SAM system continues to experience substantial drift under harsh conditions. Overall, the proposed LPVIMO-SAM demonstrates strong robustness and positioning accuracy across the collected scenarios.
	
	\begin{figure}
		\centering
		\includegraphics[width=8cm]{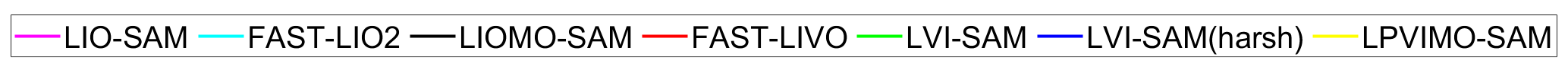}
		\includegraphics[width=3.8cm]{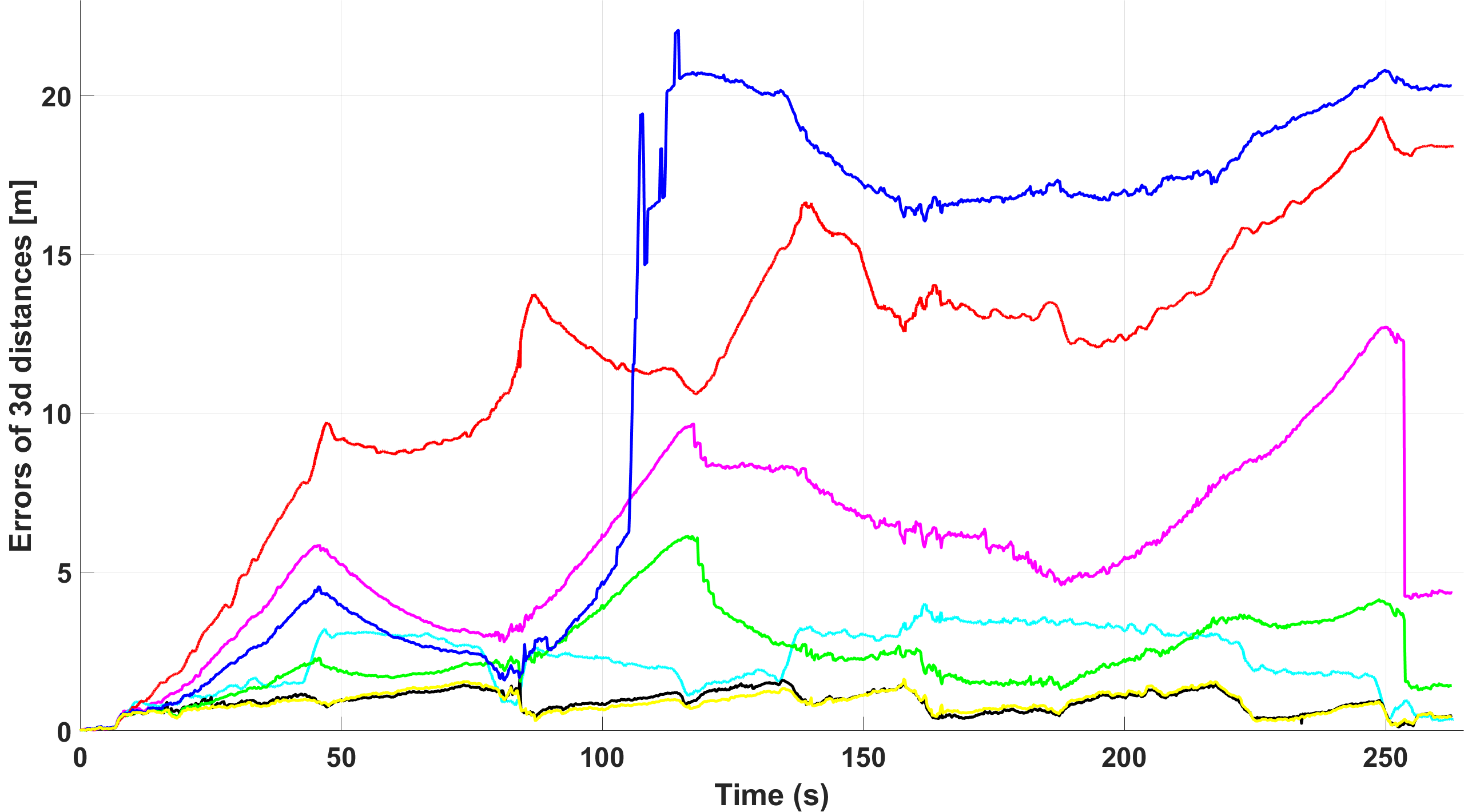}
		\includegraphics[width=3.8cm]{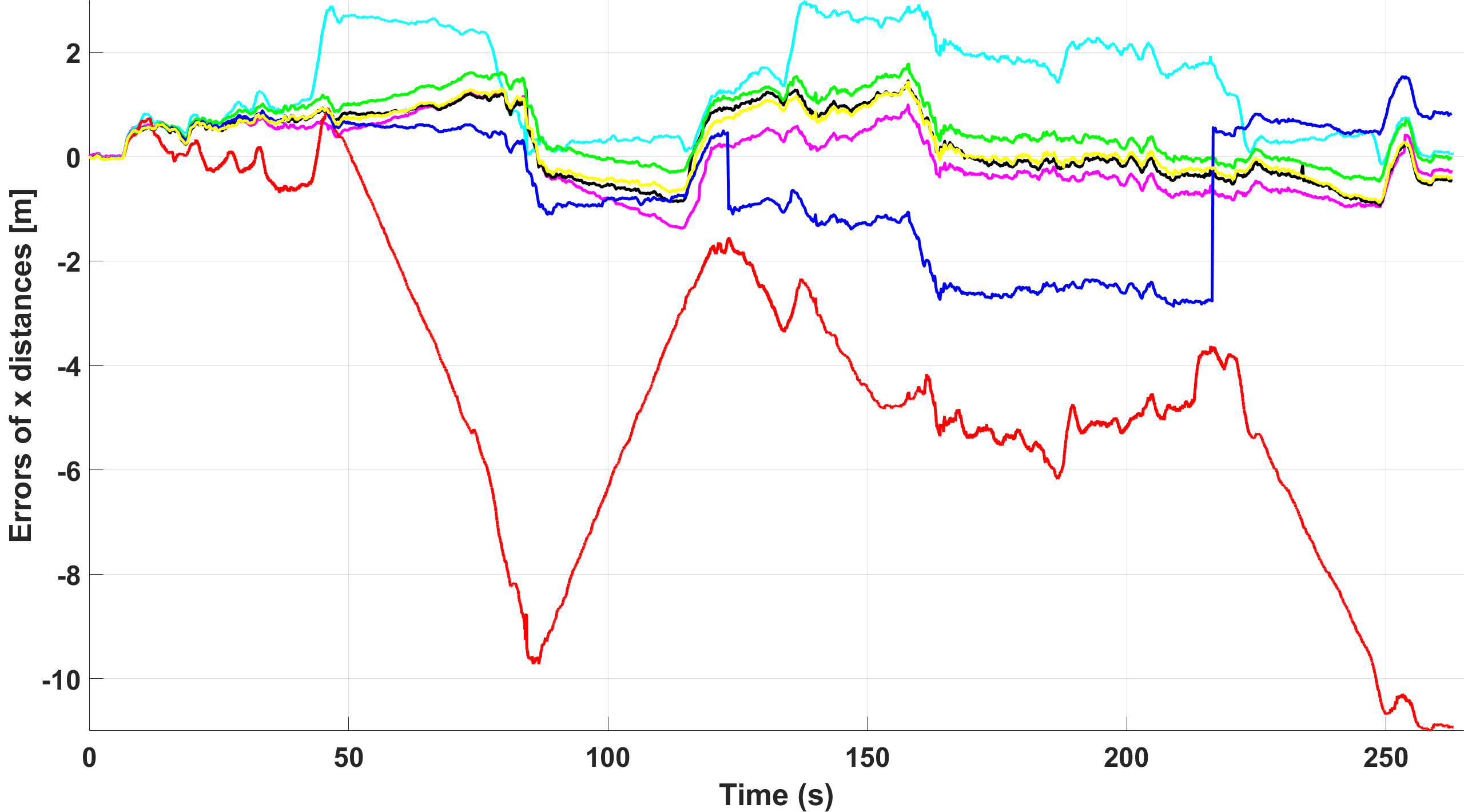}
		\\
		\includegraphics[width=3.8cm]{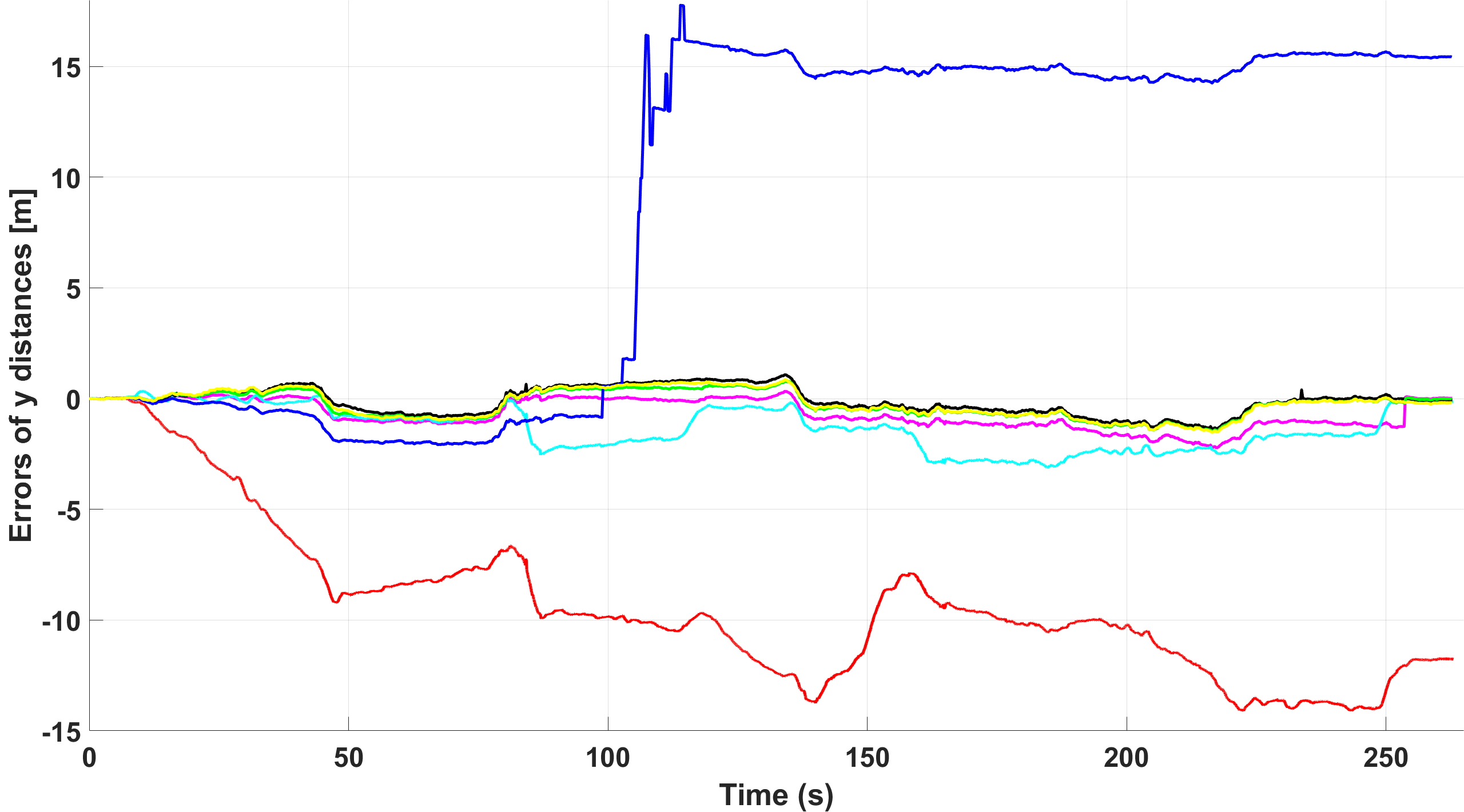}
		\includegraphics[width=3.8cm]{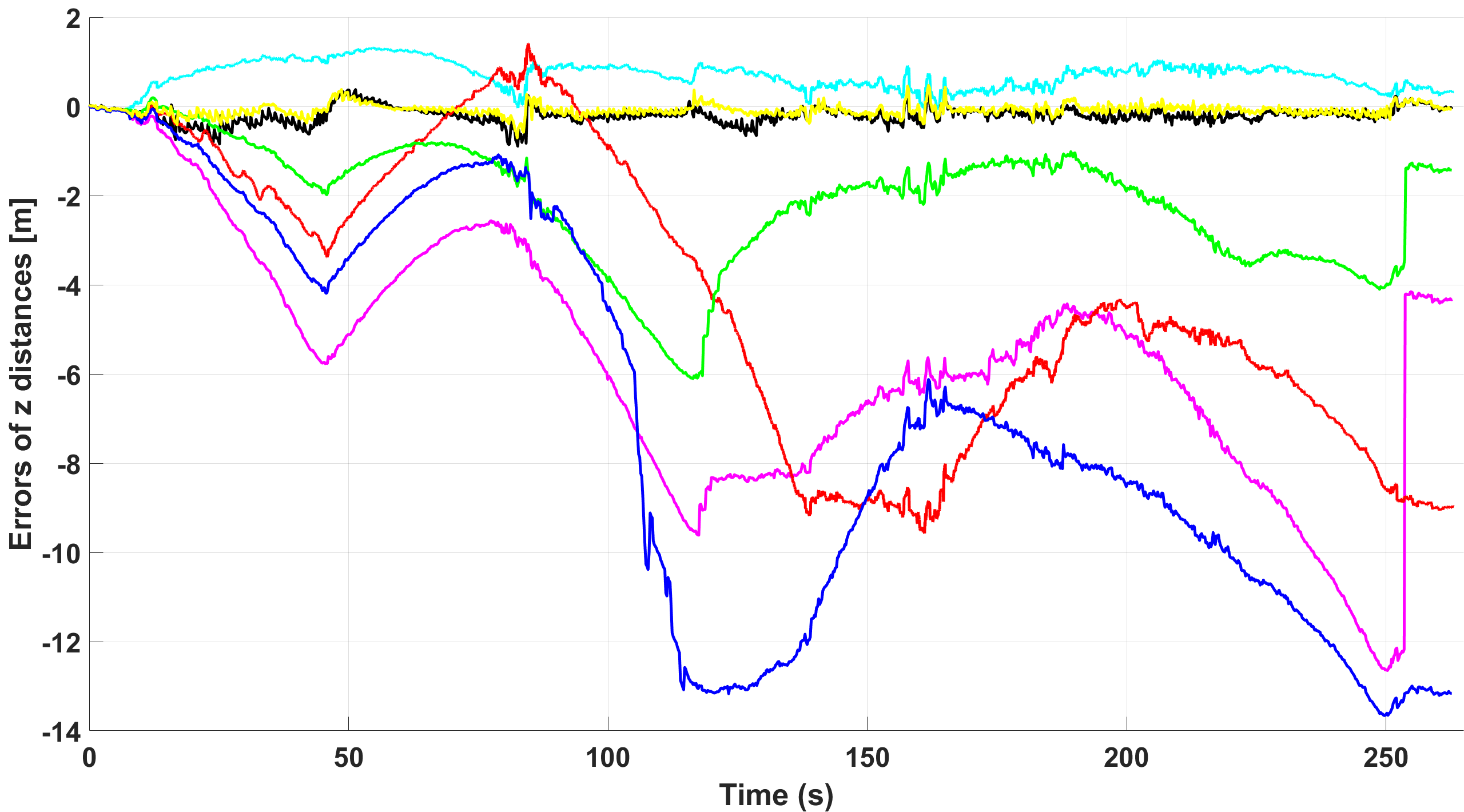}
		\caption{Comparison chart of the estimation error of the three-dimensional
			distance and the errors of the trajectory in each direction in scene d, where the 3D distance is $\sqrt{\mathrm{p}_{x}^{2}+\mathrm{p}_{y}^{2}+\mathrm{p}_{z}^{2}}$.}\label{fig5}
	\end{figure}
	To conduct a deeper evaluation of the positioning accuracy, Fig. 4 shows the trajectory comparison in scene d, and Fig. 5 presents the position estimation errors of seven methods (LIO-SAM, FAST-LIO2, LIOMO-SAM, FAST-LIVO, LVI-SAM, LVI-SAM (Harsh), LPVIMO-SAM) in this scenario. When calculating the errors, all the first nodes of the trajectories are precisely aligned with the ground truth. As shown in Fig. 5, the errors of LIO-SAM are concentrated on the Z-axis. The errors of FAST-LIO2 on the X and Y axes are higher than those of the proposed LPVIMO-SAM. The trajectory of FAST-LIVO drifts due to the influence of the selected images. Compared with LIO-SAM, LVI-SAM corrects some of the Z-axis errors, but LVI-SAM (Harsh) experiences overall drift due to insufficient features. The newly proposed LIOMO-SAM and LPVIMO-SAM perform excellently, with errors in all directions fluctuating within a small range near zero, and achieving the optimal overall trajectory positioning error.
	\begin{figure}[htbp]
		\centering
		\begin{subfigure}{4.2cm}
			\includegraphics[width=\textwidth]{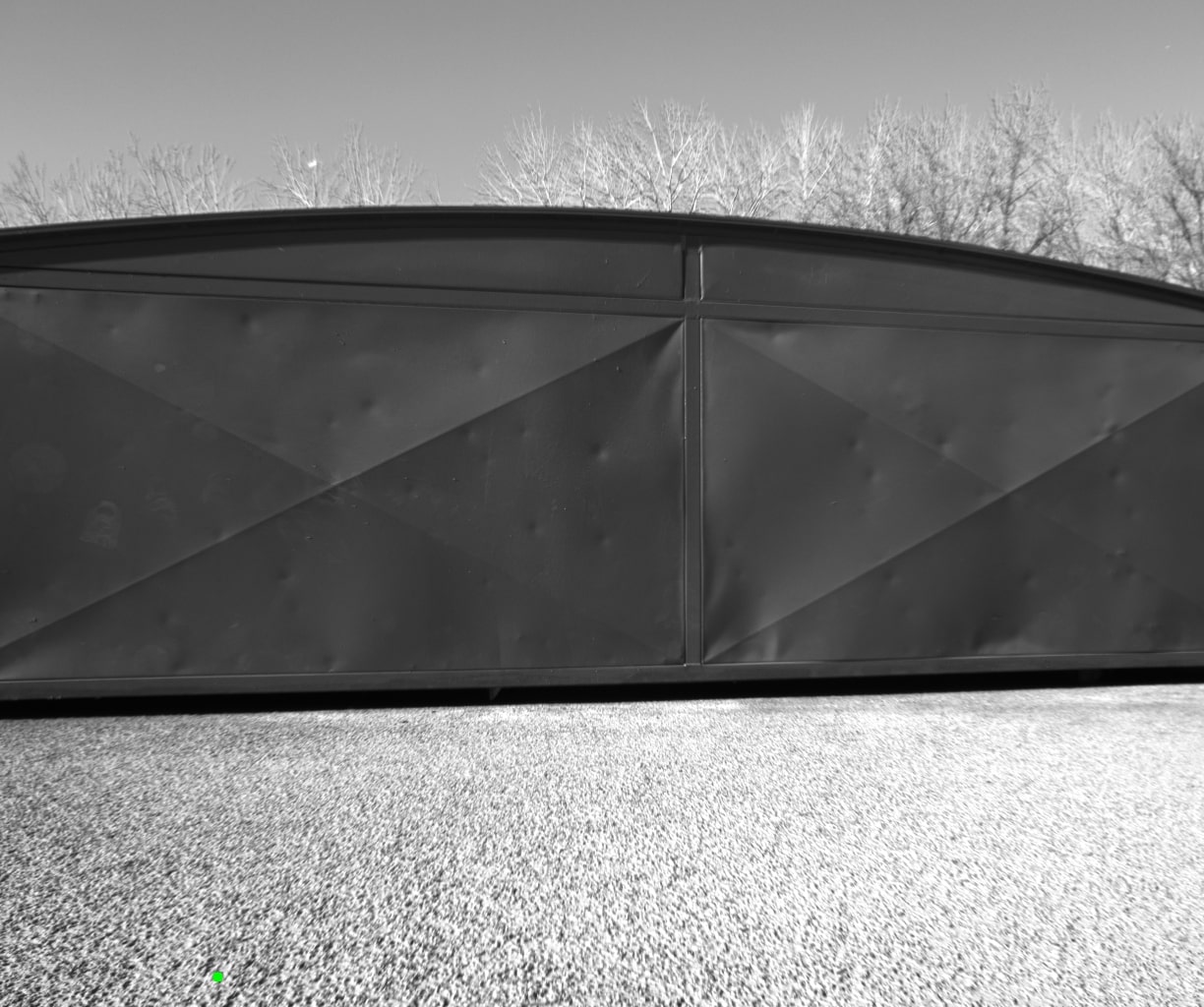}
			\caption{Grayscale graph}
		\end{subfigure}
		\hfill
		\begin{subfigure}{4.2cm}
			\includegraphics[width=\textwidth]{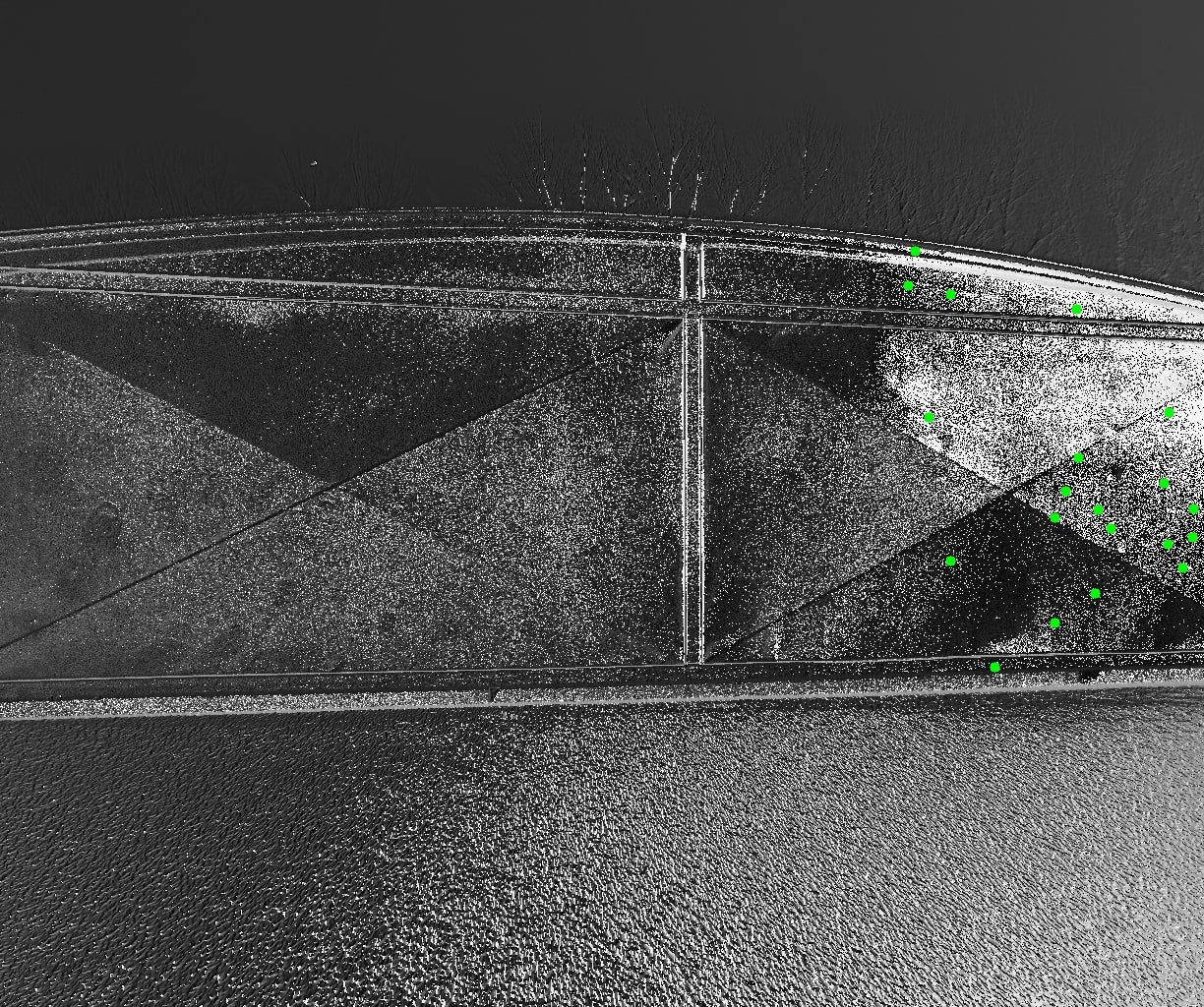}
			\caption{DOP graph}
		\end{subfigure}
		
		\begin{subfigure}{4.2cm}
			\includegraphics[width=\textwidth]{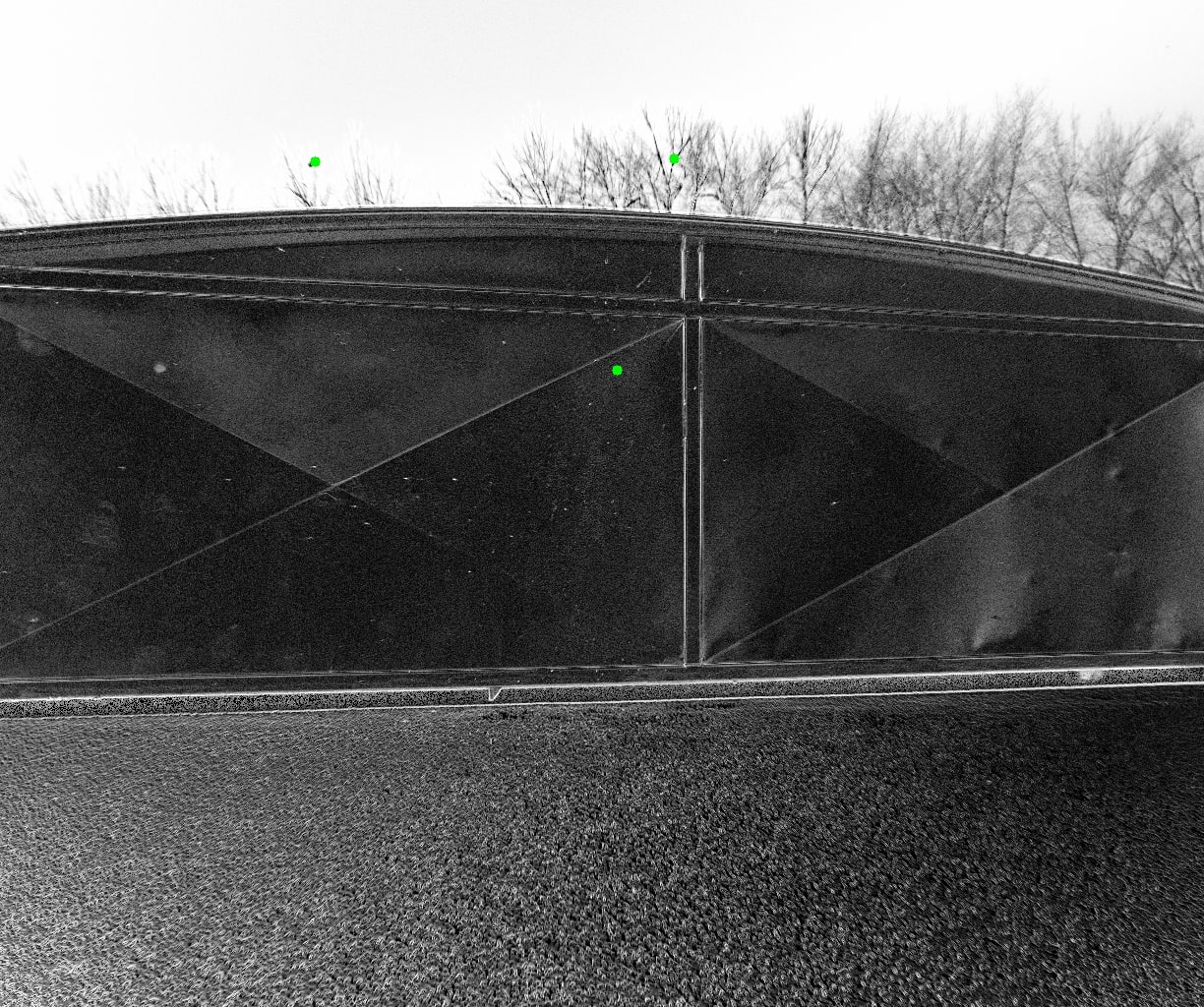}
			\caption{AOP graph}
		\end{subfigure}
		\hfill
		\begin{subfigure}{4.2cm}
			\includegraphics[width=\textwidth]{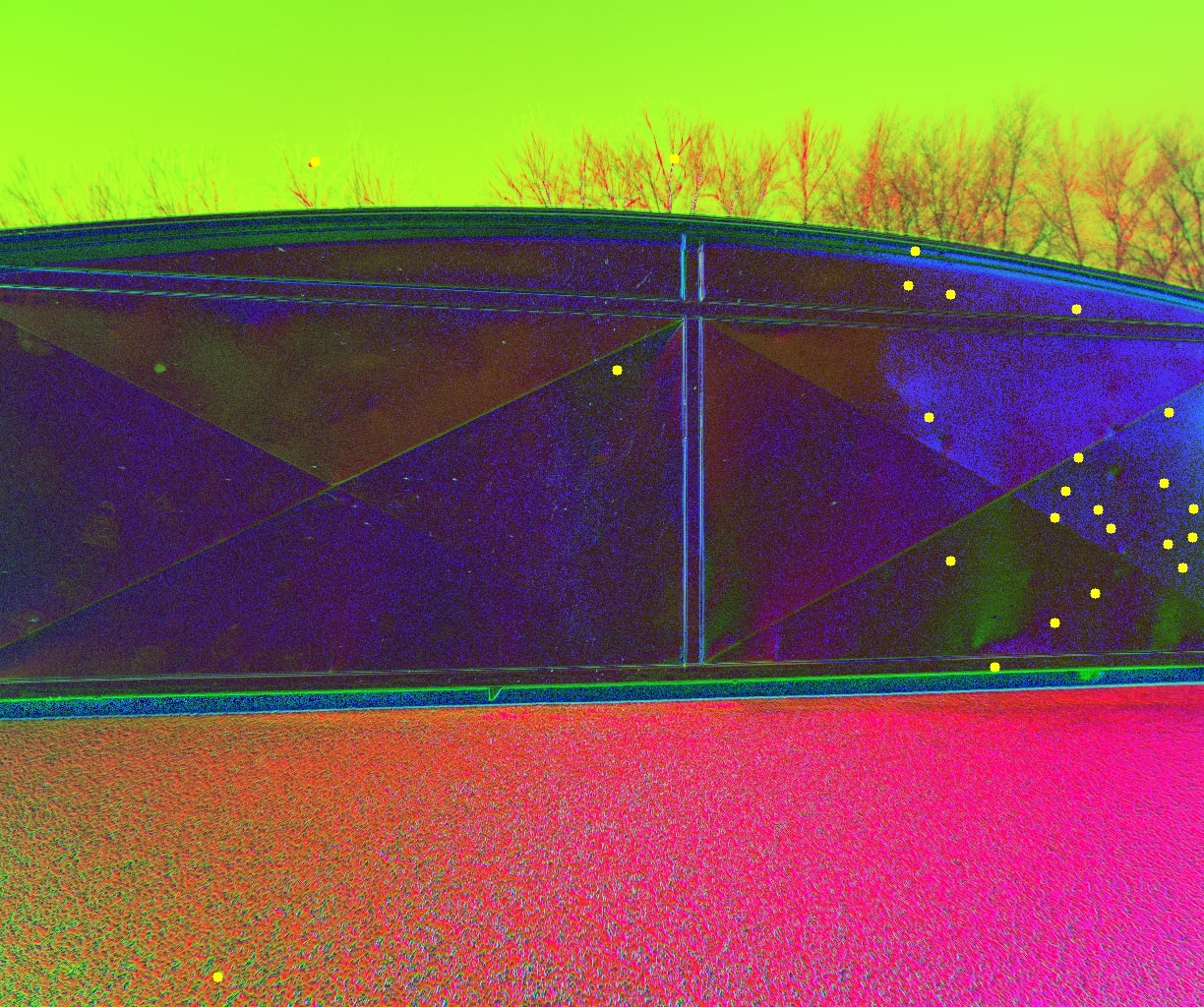}
			\caption{Polarization rgb graph}
		\end{subfigure}
		
		\caption{Image feature extraction results and polarized RGB image}\label{fig6}
	\end{figure}
	
	To fully demonstrate the feature enhancement effect of the DOP and AOP images calculated based on scene polarization information on grayscale images, we set the corner selection quality level coefficients of LVI-SAM (Harsh) and LPVIMO-SAM to 0.9-0.95 to simulate harsh conditions in low-texture scenes. As shown in Table 1 and Fig. 6, in most scenes, due to the small number of features extracted by LVI-SAM (Harsh), the VIS error is significant, causing the system to drift and even resulting in complete failure of trajectory positioning. In contrast, as shown in Fig. 6, by introducing scene polarization information in this paper and effectively extracting polarization features using the DOP and AOP images, the LPVIMO-SAM system can operate stably even in low-texture and low-feature scenes. Moreover, due to the high positioning accuracy of the LIOMO-SAM subsystem, the entire navigation system has both high stability and high positioning accuracy advantages. Its positioning accuracy is improved by approximately 36.8\% compared with that of the normally operating LVI-SAM and is higher than that of FAST-LIO2 and FAST-LIVO systems. However, the LPVIMO-SAM system is based on the factor graph architecture, and its operating speed is slower than that of the filtering-based FAST-LIO2 and FAST-LIVO systems. The number of output nodes is only about 50\% of that of FAST-LIO2 and 16.8\% of that of FAST-LIVO. Nevertheless, overall, the LPVIMO-SAM system demonstrates excellent performance in terms of positioning accuracy and stability.
	\section{Conclusion}
	For the problems of low positioning accuracy and sparse visual features in outdoor LiDAR-degraded, low-texture, and low-feature environments, we constructs a tightly-coupled LiDAR/Polarization Vision/Inertial/Magnetometer/Optical Flow integrated navigation system based on factor graph optimization. To reduce the accumulated errors from the LIO system, magnetometer is introduced to correct the heading deviation, and an optical flow ranging module is incorporated to obtain velocity and height, thereby reducing the accumulated errors and mitigating Z-axis drift. Aiming at the issue that sparse visual features in low-texture and low-feature scenes lead to visual feature matching failures in the VIS, thereby causing trajectory drift in the overall LVIS, we propose a Polarization Visual-Inertial System.  PVIS effectively enhances the robustness of VIS. Finally, we establish an error function by fusing Polarization Vision-Inertial odometry factors, IMU pre-integration factors, LiDAR odometry factors, magnetometer heading factors, loop closure detection factors, as well as the velocity and height factors from the optical flow ranging module. The optimal solution is then obtained using the iSAM2 method, further enhancing the system's positioning accuracy. Outdoor comparison experiments with LIO-SAM, FAST-LIO2, LIOMO-SAM, FAST-LIVO, and LVI-SAM are carried out, and five groups of data with different lengths and scenes are collected. The results show that the proposed LPVIMO-SAM method  exhibits good robustness and high accuracy in various environments, with an RMSE of \SI{0.87}{\metre}/\SI{256.1}{\metre} (scene d). Compared with LVI-SAM, the RMSE of the proposed LPVIMO-SAM system is reduced by  43.40\%.


\begin{thebibliography}{99}
		
		\bibitem{zhang2014loam}
		J. Zhang and S. Singh, 
		``LOAM: lidar odometry and mapping in real-time,'' 
		in \textit{Robotics: Science and Systems}, 2014, pp. 1--9.
		
		\bibitem{shan2018lego}
		T. Shan and B. Englot, 
		``LeGO-LOAM: Lightweight and Ground-Optimized Lidar Odometry and Mapping on Variable Terrain,'' 
		in \textit{IEEE/RSJ International Conference on Intelligent Robots and Systems (IROS)}, 2018, pp. 4758--4765.
		
		\bibitem{campos2021orbslam3}
		C. Campos, R. Elvira, J. J. G. Rodriguez, J. M. M. Montiel, and J. D. Tardos, 
		``ORB-SLAM3: An Accurate Open-Source Library for Visual, Visual-Inertial, and Multimap SLAM,'' 
		in \textit{IEEE Transactions on Robotics}, vol. 37, no. 6, pp. 1874--1890, 2021.
		
		\bibitem{qin2018vinsmono}
		T. Qin, P. Li, and S. Shen, 
		``VINS-Mono: A Robust and Versatile Monocular Visual-Inertial State Estimator,'' 
		in \textit{IEEE Transactions on Robotics}, vol. 34, no. 4, pp. 1004--1020, 2018.
		
		\bibitem{shan2020liosam}
		T. Shan, B. Englot, D. Meyers, W. Wang, C. Ratti, and D. Rus, 
		``LIO-SAM: Tightly-coupled Lidar Inertial Odometry via Smoothing and Mapping,'' 
		in \textit{2020 IEEE/RSJ International Conference on Intelligent Robots and Systems (IROS)}, 2020, pp. 5135--5142.
		
		\bibitem{lin2021r2live}
		J. Lin, C. Zheng, W. Xu, and F. Zhang, 
		``R2LIVE: A Robust, Real-Time, LiDAR-Inertial-Visual Tightly-Coupled State Estimator and Mapping,'' 
		in \textit{IEEE Robotics and Automation Letters}, vol. 6, no. 4, pp. 7469--7476, 2021.
		
		\bibitem{shan2021lvisam}
		T. Shan, B. Englot, C. Ratti, and D. Rus, 
		``LVI-SAM: Tightly-coupled Lidar-Visual-Inertial Odometry via Smoothing and Mapping,'' 
		in \textit{2021 IEEE International Conference on Robotics and Automation (ICRA)}, 2021, pp. 5692--5698.
		
		\bibitem{xu2023golio}
		J. Xu, Y. Li, and J. Shi, 
		``GO-LIO: Lidar-IMU SLAM Algorithm Based on Ground Optimization,'' 
		in \textit{2023 China Automation Congress (CAC)}, 2023, pp. 2809--2814.
		
		\bibitem{wang2024lowtexture}
		H. Wang, Q. Zhang, Z. Zheng, X. Li, H. Tan, and R. Li, 
		``A Low-Texture Robust Hybrid Feature Based Visual Odometry,'' 
		in \textit{2024 IEEE/RSJ International Conference on Intelligent Robots and Systems (IROS)}, 2024, pp. 10113--10120.
		
		\bibitem{shao2019stereovilidar}
		W. Shao, S. Vijayarangan, C. Li, and G. Kantor, 
		``Stereo Visual Inertial LiDAR Simultaneous Localization and Mapping,'' 
		in \textit{2019 IEEE/RSJ International Conference on Intelligent Robots and Systems (IROS)}, Macau, China, 2019, pp. 370--377.
		
		\bibitem{zhu2021camvox}
		Y. Zhu, C. Zheng, C. Yuan, X. Huang, and X. Hong, 
		``CamVox: A Low-cost and Accurate Lidar-assisted Visual SLAM System,'' 
		in \textit{2021 IEEE International Conference on Robotics and Automation (ICRA)}, 2021, pp. 5049--5055.
		
		\bibitem{huang2020lidarmonocular}
		S.-S. Huang, Z.-Y. Ma, T.-J. Mu, H. Fu, and S.-M. Hu, 
		``Lidar-Monocular Visual Odometry using Point and Line Features,'' 
		in \textit{2020 IEEE International Conference on Robotics and Automation (ICRA)}, 2020, pp. 1091--1097.
		
		\bibitem{cheng2024vehiclelocalization}
		J. Cheng, L. Zhang, Q. Chen, Z. Fu, and L. Du, 
		``High Precision and Robust Vehicle Localization Algorithm With Visual-LiDAR-IMU Fusion,'' 
		in \textit{IEEE Transactions on Vehicular Technology}, vol. 73, no. 8, pp. 11029--11043, 2024.
		
		\bibitem{mourikis2007msckf}
		A. I. Mourikis and S. I. Roumeliotis, 
		``A Multi-State Constraint Kalman Filter for Vision-aided Inertial Navigation,'' 
		in \textit{Proceedings 2007 IEEE International Conference on Robotics and Automation}, 2007, pp. 3565--3572.
		
		\bibitem{zuo2019licfusion}
		X. Zuo, P. Geneva, W. Lee, Y. Liu, and G. Huang, 
		``LIC-Fusion: LiDAR-Inertial-Camera Odometry,'' 
		in \textit{2019 IEEE/RSJ International Conference on Intelligent Robots and Systems (IROS)}, 2019, pp. 5848--5854.
		
		\bibitem{zheng2022fastlivo}
		C. Zheng, Q. Zhu, W. Xu, X. Liu, Q. Guo, and F. Zhang, 
		``FAST-LIVO: Fast and Tightly-coupled Sparse-Direct LiDAR-Inertial-Visual Odometry,'' 
		in \textit{2022 IEEE/RSJ International Conference on Intelligent Robots and Systems (IROS)}, 2022, pp. 4003--4009.
		
		\bibitem{zheng2025fastlivo2}
		C. Zheng et al., 
		``FAST-LIVO2: Fast, Direct LiDAR-Inertial-intersting taskVisual Odometry,'' 
		in \textit{IEEE Transactions on Robotics}, vol. 41, pp. 326--346, 2025.
		
		\bibitem{zhang2024lviofusion}
		H. Zhang, L. Du, S. Bao, J. Yuan, and S. Ma, 
		``LVIO-Fusion: Tightly-Coupled LiDAR-Visual-Inertial Odometry and Mapping in Degenerate Environments,''
		\textit{IEEE Robotics and Automation Letters}, vol. 9, no. 4, pp. 3783--3790, 2024.
		
		\bibitem{dupeyroux2019antbot}
		J. Dupeyroux, J. R. Serres, and S. Viollet, 
		``Antbot: A six-legged walking robot able to home like desert ants in outdoor environments,'' 
		\textit{Science Robotics}, vol. 4, no. 27, 2019.
		
		\bibitem{chiou2020neural}
		T. H. Chiou and C. W. Wang, 
		``Neural processing of linearly and circularly polarized light signal in a mantis shrimp \textit{Haptosquilla pulchella} (Miers, 1880),'' 
		\textit{Journal of Experimental Biology}, vol. 223, no. 22, p. jeb.219832, 2020.
		
		\bibitem{wang2015positionmethod}
		Y. Wang, J. Chu, R. Zhang, L. Wang, and Z. Wang, 
		``A novel autonomous real-time position method based on polarized light and geomagnetic field,'' 
		\textit{Scientific Reports}, vol. 5, p. 9725, 2015.
		
		\bibitem{yang2020polarizationahrs}
		J. Yang, T. Du, X. Liu, B. Niu, and L. Guo, 
		``Method and Implementation of a Bioinspired Polarization-Based Attitude and Heading Reference System by Integration of Polarization Compass and Inertial Sensors,'' 
		in \textit{IEEE Transactions on Industrial Electronics}, vol. 67, no. 11, pp. 9802--9812, 2020.
		
		\bibitem{dou2022polarizednavigation}
		Q. Dou, T. Du, S. Wang, J. Yang, and L. Guo, 
		``A Novel Polarized Skylight Navigation Model for Bionic Navigation With Marginalized Unscented Kalman Filter,'' 
		in \textit{IEEE Sensors Journal}, vol. 22, no. 5, pp. 4472--4483, 2022.
		
		\bibitem{du2022inslidar}
		T. Du, S. Shi, Y. Zeng, J. Yang, and L. Guo, 
		``An Integrated INS/LiDAR Odometry/Polarized Camera Pose Estimation via Factor Graph Optimization for Sparse Environment,'' 
		in \textit{IEEE Transactions on Instrumentation and Measurement}, vol. 71, pp. 1--11, 2022.
		
		\bibitem{qiu2024robustekf}
		Z. Qiu, S. Wang, P. Hu, and L. Guo, 
		``Outlier-Robust Extended Kalman Filtering for Bioinspired Integrated Navigation System,'' 
		in \textit{IEEE Transactions on Automation Science and Engineering}, vol. 21, no. 4, pp. 5881--5894, 2024.
		
		\bibitem{wan2024visualinertial}
		Z. Wan, P. Fu, K. Wang, and K. Zhao, 
		``Visual-Inertial Localization Leveraging Skylight Polarization Pattern Constraints,'' 
		in \textit{IEEE Robotics and Automation Letters}, vol. 9, no. 12, pp. 11481--11488, 2024.
		
		\bibitem{zhao2024solartracking}
		Q. Zhao, J. Yang, P. Hu, J. Qiao, A. Wang, and L. Guo, 
		``Solar-Tracking for Integrated Orientation Based on the Degree of Underwater Polarization,'' 
		in \textit{IEEE Transactions on Industrial Informatics}, vol. 20, no. 10, pp. 11521--11531, 2024.
		
		\bibitem{zhang2024headingdetermination}
		Y. Zhang, L. Guo, W. Yu, T. Chen, and S. Fang, 
		``Heading Determination of Bionic Polarization Sensor Based on Night Composite Light Field,'' 
		in \textit{IEEE Sensors Journal}, vol. 24, no. 1, pp. 909--919, 2024.
		
		\bibitem{chen2024autonomouspositioning}
		T. Chen et al., 
		``An Autonomous Positioning Method Utilizing Feature Extraction From Polarized Moonlight,'' 
		in \textit{IEEE Sensors Journal}, vol. 24, no. 14, pp. 22106--22116, 2024.
		
		\bibitem{wang2017bionicorientation}
		Y. Wang, X. Hu, J. Lian, L. Zhang, and X. He, 
		``Bionic Orientation and Visual Enhancement With a Novel Polarization Camera,'' 
		in \textit{IEEE Sensors Journal}, vol. 17, no. 5, pp. 1316--1324, 2017.
		
		\bibitem{yang2018polarimetricslam}
		L. Yang, F. Tan, A. Li, Z. Cui, Y. Furukawa, and P. Tan, 
		``Polarimetric Dense Monocular SLAM,'' 
		in \textit{2018 IEEE/CVF Conference on Computer Vision and Pattern Recognition}, 2018, pp. 3857--3866.
		
		\bibitem{Du2025ORBSLAM}
		T. Du, D. Shan, L. Huang,
		``Polarization ORB-SLAM: An ORB-SLAM Systems Assisted by  Polarization Degree and Polarization Angle for Low-Texture Environments,''  submitted to \textit{IEEE/ASME Transactions on Mechatronics}.
		
		\bibitem{kaess2012isam2}
		Michael Kaess, et al., 
		``iSAM2: Incremental smoothing and mapping using the Bayes tree,'' 
		\textit{The International Journal of Robotics Research}, vol. 31, no. 2, pp. 216--235, 2012.
		
		\bibitem{Shan2025pfliosam}
		T. Du, D. Shan, Peng. Guo,
		``PFLIO-SAM: Tightly-coupled
		IMU/LiDAR/Polarization Camera/Optical Flow
		Odometry via Smoothing and Mapping,'' submitted to
		\textit{IEEE Sensors Journal}.
		
	\end{thebibliography}

\end{document}